\documentclass[11pt]{article}
\usepackage[margin=1in, centering]{geometry}
\usepackage{graphicx} 
\usepackage{ktmacros}
\usepackage{amsmath}
\usepackage{esint}
\usepackage{mathtools}
\usepackage{amssymb}
\usepackage{amsthm}
\usepackage{xcolor}
\usepackage{bbm}

\newcommand{\calA}{\mathcal{A}}

\newcommand{\calH}{\mathcal{H}}
\newcommand{\calL}{\mathcal{L}}
\newcommand{\calG}{\mathcal{G}}
\newcommand{\calX}{\mathcal{X}}

\newcommand{\calU}{\mathcal{U}}
\newcommand{\SOA}{\mathrm{SOA}}
\newcommand{\LDim}{\mathrm{LDim}}
\newcommand{\DDim}{\mathrm{DDim}}
\newcommand{\err}{\mathrm{err}}

\title{Private Learning of Littlestone Classes, Revisited}
\author{Xin Lyu\footnote{Email: xinlyu@berkeley.edu}\\ UC Berkeley}
\date{}

\begin{document}

\maketitle

\newcommand{\xin}[1]{{\color{red} Xin: #1}}

\thispagestyle{empty}

\begin{abstract}
We consider online and PAC learning of Littlestone classes subject to the constraint of approximate differential privacy. Our main result is a private learner to online-learn a Littlestone class with a mistake bound of $\tilde{O}(d^{9.5}\cdot \log(T))$ in the realizable case, where $d$ denotes the Littlestone dimension and $T$ the time horizon. This is a doubly-exponential improvement over the state-of-the-art \cite{golowich2021littlestone} and comes polynomially close to the lower bound for this task. 

The advancement is made possible by a couple of ingredients. The first is a clean and refined interpretation of the ``irreducibility'' technique from the state-of-the-art private PAC-learner for Littlestone classes \cite{ghazi2021private-learning}. Our new perspective also allows us to improve the PAC-learner of \cite{ghazi2021private-learning} and give a sample complexity upper bound of $\widetilde{O}(\frac{d^5 \log(1/\delta\beta)}{\varepsilon \alpha})$ where $\alpha$ and $\beta$ denote the accuracy and confidence of the PAC learner, respectively. This improves over \cite{ghazi2021private-learning} by factors of $\frac{d}{\alpha}$ and attains an optimal dependence on $\alpha$. 

Our algorithm uses a private sparse selection algorithm to \emph{sample} from a pool of strongly input-dependent candidates. However, unlike most previous uses of sparse selection algorithms, where one only cares about the utility of output, our algorithm requires understanding and manipulating the actual distribution from which an output is drawn. In the proof, we use a sparse version of the Exponential Mechanism from \cite{GhaziKM20}, which behaves nicely under our framework and is amenable to a very easy utility proof.
\end{abstract}

\newpage

\section{Introduction}

We continue a long line of work on the relationship between Differential Privacy \cite{dwork2006calibrating}, the PAC-learning theory \cite{valiant1984theory} and online learning \cite{littlestone1988learning}. The theory of privacy-preserving (as measured by differential privacy) machine learning \cite{kasiviswanathan2011private-learning-theory} has been a well-motivated and prosperous research area \cite{kasiviswanathan2011private-learning-theory,beimel2014bounds,BNSV15-threshold,beimel2013characterizing,feldman2014sample,kaplan2020close-exponential-gap,cohen2023optimal}. 

While the sample complexity of pure-private learner has been fairly well-understood \cite{beimel2013characterizing,feldman2014sample}, a characterization of approximate-private learning has been elusive. Nonetheless, in the past few years, it has been realized that private learnability and online learnability are closely connected. In particular, a pair of papers \cite{alon2019private,bun2020equivalence} (and their merged version \cite{alon2022equivalence}) have demonstrated that, qualitatively speaking, Littlestone dimension, which is a combinatorial complexity measure of a hypothesis class originally studied in the online learning context, characterizes private learnability. Specifically, a hypothesis class $\mathcal{H}$ is (approximately) privately learnable if and only if it has a finite Littlestone dimension. Letting $d=\LDim(\mathcal{H})$, it was shown \cite{alon2022equivalence} that $\mathcal{H}$ can be privately PAC-learned with a sample of size $2^{2^{O(d)}}$, and cannot be privately PAC-learned with a sample of size $o(\log^\star d)$\footnote{Recall that $\log^\star n$ denotes the iterated logarithm function: namely, it is the number of times one needs to take the ``log'' of $n$ to make it smaller than $1$.}. While the gap between the upper and lower bound is by no means small, this result is exciting as it is the first to characterize the private learnability of an arbitrary hypothesis class, and this is done by connecting to the well-established area of online learning. 

The exciting breakthrough of \cite{alon2019private,bun2020equivalence} has inspired many follow-up questions, and has triggered an extensive body of research (e.g., \cite{jung2020equivalence,ghazi2021private-learning,golowich2021littlestone,bun2024private,bun2023stability,fioravanti2024ramsey}). Among them, most relevant to us are the work of Ghazi et al~\cite{ghazi2021private-learning} and the work of Golowich and Livni \cite{golowich2021littlestone}, which strengthen and tighten the connection between private and online learning from two different aspects. Namely, the work \cite{ghazi2021private-learning} quantitatively improves over the algorithm of \cite{bun2020equivalence} by giving a private PAC learner for Littlestone classes with a sample complexity of $\widetilde{O}(d^6)$. The work \cite{golowich2021littlestone} strengthens the connection by proving that every Littlestone class can be privately \emph{online} learned with a mistake bound of $2^{2^{O(d)}}$, on any realizable sequence of queries. It has been asked in \cite{golowich2021littlestone} whether one can use ideas from \cite{ghazi2021private-learning} to obtain a mistake bound of $\mathrm{poly}(d)$, and whether one can design a learner for the \emph{agnostic} case of online learning (\cite{ben2009agnostic}).

\paragraph*{Online learning versus prediction.} Apart from the pure theoretic interest in understanding the connection between online and private learning (and more broadly algorithmic stability \cite{bun2023stability}). Private online learning is, by itself, an interesting and important research direction for privacy-preserving machine learning and data analysis. For just a few examples, real-time navigation and routing (e.g.~Google Maps), digital advertisement (e.g.,~Google Ad, Amazon shopping promotion), social media and content feeding (e.g.,TikTok, YouTube short) constantly improve their service by collecting and analyzing responses and behaviors from users in an online manner. As individual privacy becomes an increasingly critical concern, it is pivotal to devise privacy-preserving principles and techniques for online learning and optimization.

Partly motivated by the lack of efficient private online learner and strong statistical barriers to private PAC learning \cite{alon2019private}, some recent works (e.g., \cite{naor2023private-everlasting-prediction,kaplan2023black}) turned attention to a relaxed privacy-preserving learning model, henceforth termed \emph{private online prediction}. In the online prediction model, one can either make distributional assumptions on the input \cite{naor2023private-everlasting-prediction} or not \cite{kaplan2023black}. The relaxation lies in the fact that the algorithm is only required to output its prediction/classification for the next query $x_{t}$, as opposed to putting forward a complete hypothesis $h_t:\mathcal{X}\to \{0,1\}$ in the standard PAC/online learning model. As such, the output of the algorithm at the $t$-th query can depend strongly on the $t$-th input, and the privacy requirement is that the algorithm's output on every other query cannot depend too strongly on $x_t$. This is formalized under the joint-differential privacy (JDP) framework. We refer interested readers to \cite{naor2023private-everlasting-prediction,kaplan2020close-exponential-gap} for more details.

The relaxation allows \cite{naor2023private-everlasting-prediction} to bypass the private PAC learning lower bound, and (under a distributional assumption) devise a private prediction algorithm whose sample complexity only scales with the (square of) VC dimension of the hypothesis. This is provably more advantageous as private PAC learning is subject to a lower bound in terms of Littlestone dimension \cite{alon2019private}. In the distribution-free setting, the work of \cite{kaplan2023black} gave an online predictor whose mistake bound is $\widetilde{O}(\log(T) d^2/\varepsilon^2)$ where $d$ denotes the Littlestone dimension. However, it is not clear whether this bound is achievable in the stronger online learning model.

\paragraph*{The gap between learning and prediction?} Intuitively it appears that a complete hypothesis is more informative and useful than a prediction made on a (private) query. Though it is conceivable that publishing a model would require much more samples than merely making predictions. On one hand, as we will see in the next section, our result gives a private online learner in the standard model (where every time step the complete description of a hypothesis $h$ is published) with a mistake bound of $\mathrm{poly}(d,\log(T))$. Thus, in terms of the dependence on $d$, the online learning model can be weaker than the online prediction model by at most a polynomial factor. On the other hand, as has been shown recently in \cite{CLNSS2024onlinethreshold, dmitriev2024growth, li2024limits}, a dependence on $\log(T)$ is \emph{necessary} for online learning, where $T$ denotes the time horizon. This is a cost that can be avoided in the online prediction model, as demonstrated by \cite{CLNSS2024onlinethreshold}. While a gap of $\log(T)$ may seem minor for most use cases, it can be a significant cost when we consider more and more fine-grained discretizations of time: for example, for applications such as search trend or recommendation system, one may want the system to update (or have the flexibility to update) its model after every \emph{second} or even sooner, while interesting updates may be sparsely distributed across the day. Understanding the applicability and differences between the two models of online learning and prediction, as well as determining the exact gap between them are both interesting open problems.

\subsection{Our Contributions}

Our results affirmatively answer the open question from \cite{golowich2021littlestone}, as described below.

\paragraph*{Characterizing private online learning up to a polynomial factor.} Our main result is a private learner for Littlestone classes with a mistake bound of $\mathrm{poly}(d,\log(T))$ in the \emph{realizable} setting, against an \emph{oblivious} adversary\footnote{This refers to an adversary who must commit to the realizable query sequence $((x_1,y_1),\dots, (x_T, y_T))$ before the algorithm starts, whereas an adaptive adversary can generate new pairs $(x_t, y_y)$ on the fly, based on the past responses of the learning algorithm, so long as the realizable condition is observed in hindsight.}.

\begin{theorem}\label{thm: intro realizable learner}
    Let $\mathcal{H}\subseteq \{h:\mathcal{X}\to \{0,1\}\}$ be a hypothesis class of Littlestone dimension $d$. Then, for every $\eps, \delta > 0$, there exists an $(\eps,\delta)$-DP online learner $\mathcal{A}$ for $\mathcal{H}$ with the following guarantee: on every fixed \emph{realizable} sequence of $T$ queries $(x_1,y_1),\dots, (x_T, y_T)$, with probability $1-\frac{1}{T^{10}}$, $\mathcal{A}$ makes at most $\widetilde{O}\left( \frac{d^{9.5} \log(T) \log(1/\delta)}{ \varepsilon } \right)$ mistakes\footnote{Throughout this paper, $\widetilde{O}(f)$ means $O(f\log^c(f))$ for an absolute contact $c > 0$.}. The privacy guarantee is with respect to the change of any single query and holds for non-realizable inputs as well.
\end{theorem}

In light of the known lower bounds of $d$ and $\Omega(\log(T))$ (\cite{littlestone1988learning,CLNSS2024onlinethreshold,dmitriev2024growth,li2024limits}) our bound characterizes the private online learning rate of Littlestone classes up to a polynomial factor. 

\paragraph*{Discussions and new questions.} We see no reason to believe that our rate is optimal. We left it as an interesting open problem to investigate the tight (up to insignificant factors) achievable mistake bound for private online learning. It is also interesting to investigate the setting of \emph{adaptive adversaries}, which can sometimes exhibit surprising separations in the privacy context \cite{asi2023private}. Currently it seems some adaptation of our techniques should give a $2^{O(d)}$ bound (see the remark following Lemma~\ref{lemma: bounded mistakes}), but this is far from satisfactory: we ask if it can have a companion lower bound, or there are better upper bounds achievable.

Note that if a strong lower bound result should be discovered against adaptive adversaries, it would make \emph{two} very intriguing separations: (1) one between online learning vs.~prediction, where for the latter we have efficient algorithms \cite{kaplan2023black}, and (2) the other about online learning against an oblivious vs.~adaptive adversary. As a starting point to this investigation, we suggest it may be instructive to understand whether the idea of ``embedding multiple one-way marginal queries in a stream'' from \cite{JainRSS23} can be adapted to this setting.

\paragraph*{Improved private PAC Learning.} The starting point of our online learner is the private PAC learner from \cite{ghazi2021private-learning} in the batch setting. In particular, we give a clean and refined re-interpretation of the ``irreducibility'' notion of \cite{ghazi2021private-learning}. Our new perspective allows us to improve the sample complexity of private PAC learning as well.

\begin{theorem}
Let $\calH$ be a class of Littlestone dimension $d$. For any $\eps,\delta, \alpha, \beta$, there is an $(\eps,\delta)$-DP algorithm that PAC-learns $\calH$ up to error $\alpha$ with confidence $1-\beta$, using $\widetilde{O}\left( \frac{\log(1/\delta) d^5}{\eps \alpha} \right)$ examples.
\end{theorem}

Compared with the $\widetilde{O}\left( \frac{\log(1/\delta) d^6}{\eps \alpha^2} \right)$ bound of \cite{ghazi2021private-learning}, we are able to shave a factor of $d$ from the upper bound. More importantly, our bound obtains an optimal dependence on the accuracy parameter, $\alpha$. We do not believe our rate is optimal, but we do hope the ideas developed in our paper can help further progress on this important question.

\paragraph{Technical contributions.} On a technical level, we highlight our use of the private sparse selection technique (see, e.g, \cite{BNS13-private-learning,BNSV15-threshold, GhaziKM20}). In many previous uses of private selection, one cares about the ``utility'' of the output, and proves that the output is sub-optimal than the best candidate by at most a $\log(\#\text{candidates})$ factor. In our analysis, we crucially analyze and exploit the \emph{distribution} from which the private algorithm draws its output. Namely, it is important for us that the private selection algorithm samples (approximately uniformly) from all reasonably-good candidates. We find this algorithmic idea and analysis quite interesting and rarely seen in the ``sparse selection'' context.


Our proofs also develop some new algorithmic ideas and privacy-preserving principles that are specific to the task of learning Littlestone and VC classes. We detail them in Section~\ref{sec: technique overview}.

\section{Technique Overview}\label{sec: technique overview}

In this section, we discuss some of the key proof ideas behind our result.

\paragraph*{Interleaving hypothesis classes from split-and-aggregate.} A simple but remarkably powerful scheme we learned from \cite{ghazi2021private-learning} is the following. Say we are given an input of $N$ examples $S = (x_1,y_1),\dots, (x_N, y_N)$. We use one of the most popular private algorithm design strategies, namely split-and-aggregate, to randomly split the examples into $k$ chunks, each of size $\frac{N}{k}$, denoted by $S_1,\dots, S_k$. For a hypothesis $h$ and a sub-data set $S_i$, we may evaluate
\begin{align*}
    \err_{S_i}(h) := \E_{(x,y)\sim S_i}[h(x)\ne y].
\end{align*}
Let $\alpha = \err_S(h)$. By applying a multiplicative Chernoff bound, we find that with high probability (i.e.,~$1 - 2^{-\Omega(\frac{N}{k}\cdot \frac{\alpha}{d^2})}$), the following is true: 
\begin{align*}
    \err_{S_i}(h) \in (1\pm \frac{1}{5d}) \err_S(h).
\end{align*}
This implies the following happen with high probability:
\begin{align*}
    (1-\frac{1}{3d})\err_{S_{i'}}(h) \le \err_{S_i}(h) \le (1-\frac{1}{3d})^{-1} \err_{S_{i'}}(h).
\end{align*}
Setting $N$ larger by a factor of $d$, we can afford to union-bound over all relevant $h$ from a VC class $\mathcal{H}$ (there are at most $N^d$ such hypotheses by Sauer's lemma), and arrive at the following corollary: define hypothesis classes $H_{i}^j:= \{ h\in \mathcal{H}: \err_{S_i}(h) \le (1-\frac{1}{3d})^{j} \alpha \}$ for $i\in [k]$ and $j\le d$. Then we have, between every two groups $i$ and $i'$:
\begin{align*}
    H_{i}^{j} \subseteq H_{i'}^{j-1} \subseteq H_{i}^{j-2} \subseteq H_{i'}^{j-3} \subseteq \dots.
\end{align*}
Or, we find the following formulation easier to work with: for every $i,j$, we have
\begin{align}
    H_{i}^{j}\subseteq \bigcap_{i'} H_{i'}^{j-1}. \label{equ: Hij in cap of Hij-1}
\end{align}
This suggests the following intuitive roadmap: we initialize $k$ non-private ``teacher'' algorithms with the classes $H_1^1,\dots, H_{k}^1$, and try to train a model by running a certain aggregation procedure among $H_1^1,\dots, H_{k}^1$. This step can be done in a private manner because $H_i^1$'s were constructed from disjoint inputs. If the aggregation succeeds in outputting a ``stable'' model, the task is done. Otherwise, intuitively it is the case that difference $H_i^1$'s are capturing ``different aspects'' of the problem. In this case, we let each of the $k$ teachers pass to $H_{i}^2$. By doing so, we guarantee that each teacher is granted the collective knowledge of all teachers from the last stage, in the sense of Eq.~\eqref{equ: Hij in cap of Hij-1}. We expect a fast learning progress through this operation. As it turns out, we can devise the whole training algorithm with at most $d$ such stages of training, and hence the parameter range of $j\le d+1$.

\paragraph*{Comparing with \cite{ghazi2021private-learning}.} We conclude this part by comparing our implementation of the interleaving scheme with the original one of \cite{ghazi2021private-learning}. Our implementation differs in two aspects: first our interleaving properties are stated between pairs of teachers (or, between each teacher and the complete data set), with no reference to the underlying distribution $\mathcal{D}$, whereas \cite{ghazi2021private-learning}'s design was always with reference to the distribution $\mathcal{D}$ from which the data set is drawn. This extension makes the technique more convenient in online algorithm design: in the distribution-free online setting, there is no such ``underlying distribution'' anyway. Second and perhaps more importantly, we observe that one can use multiplicative concentration inequality to construct the scheme with fewer samples (this is ultimately the reason we can shave off a $\frac{1}{\alpha}$ factor from the PAC learner upper bound).

\paragraph*{Irreducibility of Littlestone classes.} We briefly discuss the irreducibility of Littlestone classes, first introduced by \cite{ghazi2021private-learning}. From our perspective, the irreducibility notion tries to capture the following simple phenomenon: say $\mathcal{H}$ and $\mathcal{G}$ are two hypothesis classes of the same Littlestone dimension $d$. It could be the case that $\mathcal{H}\cap \mathcal{G}$ has dimension $d$ as well: that is, by passing to $\mathcal{H}\cap \mathcal{G}$, we are not making any progress. This can be the case even if the standard optimal algorithms (SOA, see Section~\ref{sec: decomposition} for a quick review of its definition) of $\mathcal{H}$ and $\mathcal{G}$ are different! Still, suppose $\SOA_{\mathcal{H}}\ne \SOA_{\mathcal{G}}$ at a point $x$. Then, it must be the case that the classes $\{ h\in \mathcal{H}\cap \mathcal{G} : h(x) = b\}$ have strictly smaller Littlestone dimension for both $b\in \{0,1\}$. That is, $\mathcal{H}\cap \mathcal{G}$ is reducible in the following sense: by making one additional query $x$ to $\mathcal{H}\cap \mathcal{G}$, we guarantee that the Littlestone dimension of the restricted class is \emph{reduced}, hence simplifying the problem!

We refer the readers to Section~\ref{sec: decomposition} for the formal development of the technique. In particular, Lemma~\ref{lemma: p-decomposition} and the following discussion therein mirrors our description of interleaving hypothesis classes above. We also explain there how our formulation of the technique allows us to improve the sample complexity of the PAC learner from $d^6$ to $d^5$.

\paragraph*{Split-and-aggregate with online inputs.} We have reviewed the two key ideas in the previous algorithm \cite{ghazi2021private-learning}. Next, we describe our strategy to convert them into online learners. Our simple observation is that an existing interleaving scheme $\{H_i^j\}_{i\in [k], j\in [d+1]}$ can be extended to accommodate for an ensemble of extra training examples $S$. To do so, one just constructs an interleaving scheme from $S$, denoted by $\{{H'}_{i}^{j}\}_{i\in [k],j\in [d+1]}$, and takes the entry-wise intersection between $H_{i}^j$ and ${H'}_i^j$. It is straightforwardly verified that the resulting hypotheses collection is a valid scheme. 

This suggests a natural approach for online learning in the realizable setting: train $k\approx \poly(d, \log(T))$ ``teachers'', initially with \emph{empty} data sets. Then, use the output model $\hat{h}$ to answer the online queries to come. Once there are enough number of ``counter-examples'' to $\hat{h}$, collect all of them and construct an interleaving scheme $\{H_i^j\}$ from it. Privately train the $k$ teachers with the addition of the new scheme $\{H_i^j\}$ and publish a new hypothesis. The hope of doing so is that the teachers will gradually reduce the space of candidate hypotheses, end up with the only consist hypothesis and make no more mistakes.

\paragraph*{Ruling out hypotheses efficiently.} Implementing the general strategy above needs care. Here we describe what we think the most intriguing and challenging issue in our algorithm design and analysis: under the ``interleaving hypotheses'' and ``irreducibility'' framework, we will have a total of $(d+1)$ stages in the online learning process. During each stage, there can be possibly $2^{O(d^2)}$ many ``alive'' hypotheses, which are \emph{strongly} input dependent, and any of them is a good hypothesis to publish (think of these as hypotheses that are \emph{approximately} consistent with the examples seen so far). Say we choose one $\hat{h}$ from them and publish it. We wait until $\hat{h}$ makes $\mathrm{poly}(d)$ mistakes, and we update the teachers with these ``counter examples''. By doing so, we guarantee that $\hat{h}$ is no longer a viable option. Then, a pessimist may ask what happens with the other alive hypotheses: what if none of the remaining $2^{O(d^2)}-1$ alive hypotheses are affected by the counter examples? Do we need to repeat the same arguments for as many as $2^{O(d^2)}$ times, and end up with a mistake bound of $2^{O(d^2)}$?

In fact, we can run this naive argument and end up with a $2^{O(d^2)}$ bound, which we note is already an improvement to \cite{golowich2021littlestone}'s $2^{2^{O(d)}}$ bound. However, there are more efficient solutions. Recall how such issues can be resolved in the non-private version: simply use the halving algorithm: namely take the majority vote of all alive candidates. Every time the majority vote fails, at least half of all alive candidates are ruled out! Taking the majority vote over all alive hypotheses seems too much in the private context: taking an analogy with something well-known in privacy literature: \emph{selecting} one hypothesis (``learning'') is usually much easier than \emph{aggregating} all hypotheses (``sanitation''), see e.g. \cite[Section 6.2]{ghazi2021private-learning}.

\paragraph*{Oblivious stream and uniform convergence.} Since we are designing algorithms against an oblivious adversary, we can fix one such sequence $(x_1,\dots, x_T)$ before our algorithm starts. Now, suppose we have a distribution $\mathcal{D}$ over the alive hypotheses to draw from (ideally the uniform distribution), we can draw $O(\log(T))$ hypotheses from the distribution, with high probability their majority vote approximates the majority vote of $\mathcal{D}$, at least on the fixed query sequence $(x_1,y_1),\dots, (x_T, y_T)$ (this is the only place we utilize the oblivious adversary assumption). Plus, drawing $O(\log(T))$ hypotheses bring only an $O(\log(T))$ overhead in terms of privacy cost.

This is where we find, conveniently and somewhat surprisingly, that the exponential mechanism (which is originated from \cite{McSherryT07}, and we use a sparse variant of it from \cite{GhaziKM20}), gives exactly what we ask for: it ensures privacy of the algorithm, and samples a candidate approximately uniformly from the pool of all hypotheses, assuming they are equally good. Moreover, the explicit description of the distribution (namely $\frac{\exp(v_i)}{\int \exp(v_{i'})}$) makes it \emph{very clear} what happens when we rule out a subset of hypotheses (in contrast, we are currently not able to carry out the analysis using other private selection protocol, such as Report-Noisy-Max with Laplace or Gaussian noises). It appears that our result gives the first example about how to use Exponential Mechanism to implement a version of the ``halving'' algorithm for an online task.


\paragraph*{Remark: exponential mechanism and online learning.} We should make it clear that we are by no means the first to observe that Exponential Mechanisms are ``compatible'' with online learning (the exponential mechanism can actually be seen as an instantiation of multiplicative weights): see e.g., \cite{jain2012dp-online-learning,agarwal2017price,asi2023private} for a related line of works on private online \emph{expert} problem and online convex optimization. Still, as far as we know, our result provides the first example where a private \emph{sparse} selection algorithm is understood as a sampling algorithm. Due to the ``sparsity'' nature of the problem, the pool of effective candidates will be strongly input-dependent (whereas in, e.g., the expert problem, one can assign a uniform weight to every expert as a ``prior'') to avoid a $\log(\text{size of universe})$ dependence (we note the universe size in our setting can be infinite).


\section{Preliminaries}

\subsection{Concentration inequalities}

We frequently use the following \emph{multiplicative} version of the Chernoff bound.

\begin{proposition}\label{prop:multiplicative-chernoff}
    Suppose $X_1,\dots, X_n$ are independent random variables taking values in $\{0,1\}$. Let $X=\sum_{i=1}^n X_i$ denote their sum and $\mu = \Ex[X]$. Then, for any $\delta \in (0, 1)$:
    \begin{align*}
        \Pr[|X - \mu| > \delta \mu ] \le \exp(-\frac{\delta^2\mu}{3}).
    \end{align*}
\end{proposition}

There is also a sample-without-replacement version of it, which is what being actually used in our algorithm.

\begin{proposition} \label{prop:chernoff-sample-without-replacement}
    Let $X_1,\dots, X_N\in \{0,1\}$ be $N$ integers. For $k\le N$ and $\delta \in (0, 1)$, let $S\subseteq[N]$ be a random subset of size $t$,
    \begin{align*}
        \Pr\left[ \left| \E_{i\sim S}[X_i] - \E_{i\in [N]}[X_i] \right| >  \delta \E_{i\in[N]}[X_i] \right] \le 2\exp\left(-\frac{\delta^2 t\E_{i\in [N]}[X_i]}{3}\right).
    \end{align*}
\end{proposition}


\subsection{Differential Privacy}

We assume basic familiarities with Differential Privacy, specifically its definition, the composition properties of Differential Privacy, and basic private algorithms such as the Laplace noise mechanism. The textbook of Dwork and Roth \cite{DworkR14} provides an excellent reference. In the following, we review two slightly more advanced tools.

\subsection{Private Sparse Selection and Sampling}

We need a private algorithm to select (or, rather, sample) an item from an \emph{unbounded} domain $\calU$ according to certain score functions. This is known to be impossible when pure-DP is required, or when the number of items to be sampled can be enormous. However, if the number of ``relevant'' items (given the data set $D$) is very few, there are various known approaches. 

It will be most convenient for us to use the sparse selection/sampling algorithm from \cite{GhaziKM20}. We review their algorithm in Algorithm~\ref{algo:private sample}.

\begin{algorithm}[h]
\LinesNumbered
    \caption{Private Sparse Sample}
    \label{algo:private sample}
    
    \SetKwProg{Fn}{Function}{:}{}
    
    \DontPrintSemicolon
    
    \SetKwFunction{PrivateSample}{PrivateSample}
    
    \KwIn{
         Domain $\calU$; $k\ge 1$ subsets $\calL_1,\dots, \calL_k\subseteq U$; Parameter $\eps$ and $B\ge 0$.
    }
    \KwOut{
        An item $u\in \bigcup_{i=1}^k \calL_i$ \emph{or} a failure symbol $\perp$.
    }
    
    \Fn{\PrivateSample{}}{
        Define $\mathrm{score}(u) := |\{i : u\in \calL_i \}|$ for every $u\in \bigcup_{i=1}^k \calL_i$ \;
        Define $\mathrm{score}(\perp) := B$ \;
        \Return{$v$ from $\bigcup_{i=1}^k \calL_i \cup \{ \perp\}$ where $\Pr[v \text{ is returned }] \propto \exp(\eps\cdot \mathrm{score}(v))$ \;}
    }
\end{algorithm}

We call Algorithm~\ref{algo:private sample} a ``sampling'' algorithm instead of a ``selection'' algorithm, because (as already alluded in Section~\ref{sec: technique overview}), it is important for us to understand and exploit the distribution over items that Algorithm~\ref{algo:private sample} is sampling from.

Algorithm~\ref{algo:private sample} is private so long as each set $\calL_i$ is not too large (hence ``sparse'' sampling), and the parameter $B$ is set correctly. See the following lemma.

\begin{lemma}\label{lemma: privacy of sparse sampling}
    Consider Algorithm~\ref{algo:private sample}. Suppose we have the promise that each $\calL_i$ has size at most $L$. Then, provided that $B\ge \frac{10\cdot \log(L/\delta)}{\eps}$, Algorithm~\ref{algo:private sample} is $(2\eps,\delta)$-DP with respect to the addition/removal/replacement of any single $\calL_i$.
\end{lemma}

For a proof of Lemma~\ref{lemma: privacy of sparse sampling}, see \cite[Lemma 36]{GhaziKM20}. We remark that the size upper bound on $\calL_i$ can be enforced inside Algorithm~\ref{algo:private sample} by truncating $\calL_i$ if needed. So, we can always guarantee the privacy of Algorithm~\ref{algo:private sample}. But the truncation may compromise the utility of the algorithm.




\subsection{The AboveThreshold Algorithm}

We also need the well-known AboveThreshold algorithm (a.k.a.~the Sparse Vector Technique) from the literature. Roughly speaking, the AboveThreshold algorithm allows one to privately process a sequence of sensitivity-$1$ queries $f_1,f_2,\dots, $ while only reporting (and paying for privacy) for those queries that have an evaluation larger than a pre-determined threshold $H$. We use the Cohen-Lyu \cite{CohenL23} implementation of AboveThreshold for its simplicity. The algorithm is described in Algorithm~\ref{algo: abovethreshold}.

\begin{algorithm}[h]
\LinesNumbered
    \caption{The AboveThreshold Algoirthm, \cite{CohenL23} style}
    \label{algo: abovethreshold}
    
    \SetKwProg{Fn}{Function}{:}{}
    
    \DontPrintSemicolon
    
    \SetKwFunction{AboveThreshold}{CLAboveThreshold}
    
    \KwIn{
         Private data set $D$, threshold $H\in \mathbb{R}$. Parameters $\eps > 0$ and $K\in \mathbb{N}$ \;
    }
    
    \Fn{\AboveThreshold{}}{
        $\mathrm{counter} \gets 0$ \;
        \While{$\mathrm{counter} \le K$}{
            Receive the next query $f_t$\;
            \eIf{$f_t(D) + \mathrm{Lap}(1/\eps) \ge H$}{
                Report ``Above'' \;
                $\mathrm{counter} \gets \mathrm{counter} + 1$ \;
            }{
                Report ``Below'' \;
            }
            $t\gets t + 1$ \;
        }
    }
\end{algorithm}

The advantage of \cite{CohenL23} is that there is no need to add noise to the ``threshold'' (as was done in the more standard implementation \cite{DworkR14}). Moreover, the ``above-threshold'' test is independently performed for every query by a simple Laplace noise mechanism, rendering a very straightforward utility analysis. The downside, however, is that the implementation only gives \emph{approximate}-DP guarantee, and incurs an additional \emph{additive} $O(\log(1/\delta)\cdot \eps)$ privacy cost on ``epsilon''. However, as it turns out, this additive overhead is not the bottleneck of our algorithm anyway.

We state the privacy property of Algorithm~\ref{algo: abovethreshold} below.

\begin{lemma}\label{lemma: privacy of abovethreshold}
Assuming the queries $f_t$ sent to Algorithm~\ref{algo: abovethreshold} all have sensitivity at most $1$, for every $\delta > 0$, Algorithm~\ref{algo: abovethreshold} is $(\eps \cdot O(\sqrt{K\log(1/\delta)} + \log(1/\delta)),\delta)$-DP
\end{lemma}

Two remarks are in order about \Cref{algo: abovethreshold}. First, instead of always incrementing counter for ``Above'' outcomes, we can choose to increment for all  `Below'' outcomes, and halt the algorithm as soon as $K$ ``Below''s are observed. \Cref{lemma: privacy of abovethreshold} will hold for the new algorithm just identically: to see this, simply note that in \Cref{algo: abovethreshold}, the cases of ``Above'' and ``Below'' are completely symmetric. Secondly, in our algorithm design, the queries made to \Cref{algo: abovethreshold} will be \emph{interleaved} with other private mechanisms. As such, a priori we shouldn't analyze the privacy of \Cref{algo: abovethreshold} as a standalone part and compose its privacy guarantee with other components as if we were running these components sequentially. Fortunately, the \emph{concurrent} composition theorem of differential privacy \cite{VadhanW21,Lyu22,Vadhan023} tell us that we can do so: namely we can obtain the final privacy guarantee of the whole algorithm, by using an analysis where we compose \Cref{algo: abovethreshold} with other components sequentially.

\section{Irreducibility and Decomposition Dimension}\label{sec: decomposition}

In this section, we refine and extend the notion of irreducibility of Littlestone classes, which first appeared in \cite{ghazi2021private-learning}.

\paragraph*{Notation.} We set up our notation and review the basic background of online learnability. Fix $\calX$ to be the input domain (which can possibly be unbounded). By an ``example'' we always mean a pair of the form $(x,y) \in \calX\times \{0, 1\}$. Let $\calH \subseteq \{ h: \calX\to \{0,1\}\}$ be a hypothesis class. The Littlestone dimension of $\calH$, denoted by $\LDim(\calH)$, is the maximum $d\in \mathbb{N}$ such that there is a complete depth-$d$ binary tree $T$ with the following property: every internal node of $T$ is labeled by a point $x\in \calX$ and has two outgoing edges labeled by $0$ and $1$. Every leaf of the tree is explained by a hypothesis $h\in \calH$: that is, every root-to-leaf path naturally corresponds to a sequence of examples $(x_1,y_1),\dots, (x_d, y_d)$, and there is at least one hypothesis $h\in \calH$ consistent with all these pairs. It is a well-known fact that the Littlestone dimension upper-bounds the VC dimension. So any generalization argument made for a VC class applies equally well to a Littlestone class (we will frequently use this fact without further notice). 

Given a hypothesis class $\calH$ and an example $(x,y)$, define the restriction class $\calH|_{(x,y)}$ as $\calH|_{(x,y)} = \{ h\in \calH: h(x) = y \}$. For a sequence of examples $(x_1,y_1),\dots, (x_k, y_k)$, the definition of $\calH|_{(x_1,y_1),\dots, (x_k, y_k)}$ is analogous. The standard optimal algorithm (SOA) of a class $\calH$ is a function $\SOA_{\calH}:\calX\to \{0, 1\}$ defined as:
\begin{align*}
    \SOA_{\calH}(x) = \begin{cases}
        0 & \text{if $\LDim(\calH|_{(x,0)}) = \LDim(\calH)$,} \\
        1 & \text{otherwise.}
    \end{cases}
\end{align*}
It is well known that for any $x\in \mathcal{X}$, one has $\min\left\{ \LDim(\calH|_{(x,0)}),\LDim(\calH|_{(x,1)}) \right\}< \LDim(\calH)$. Hence, for online learning of Littlestone classes in the mistake bound model, one canonical optimal strategy is to always respond with $\SOA_{\calH|_{(x_1,y_1),\dots, (x_{t-1},y_{t-1})}}$ where $(x_1,y_1),\dots, (x_{t-1},y_{t-1})$ denote the observed examples up to time step $t-1$. In this way, each time the strategy makes a mistake, the restricted class has its Littlestone dimension reduced by at least one. Hence, at most $d$ mistakes can be made for any \emph{realizable} query sequence.

\subsection{Irreducibility}

In this subsection, we review the notion of irreducibility.

\begin{definition}\label{def: irreducible}
    Let $k\in \mathbb{N}$. A hypothesis class $\calH$ is called $k$-irreducible, if for every sequence of $k$ points $x_1,\dots, x_k$, it holds that \begin{align*}
        \LDim(\calH|_{(x_1,\SOA_\calH(x_1)),\dots, (x_k,\SOA_{\calH}(x_k))}) = \LDim(\calH).
    \end{align*}
    Contrapositively, we say $\calH$ is $k$-reducible, if there exists $k$ points $x_1,\dots, x_k$, such that    
    \begin{align*}
        \LDim(\calH|_{(x_1,\SOA_\calH(x_1)),\dots, (x_k,\SOA_{\calH}(x_k))}) < \LDim(\calH).
    \end{align*}
\end{definition}


One interpretation of Definition~\ref{def: irreducible} is the following: being $k$-reducible means that $\calH$ can be \emph{covered} by $(k+1)$ \emph{restriction} classes $\calH_1,\dots, \calH_k, \calH_{k+1}$, such that each of these $\calH_i$ has a Littlestone dimension strictly less than $\LDim(\calH)$. Indeed, say the $k$-reducibility is witnessed by $(x_1,\dots, x_k)$. For any $i\in [k]$, we have $\LDim({\calH|_{(x_i,1 - \SOA_{\calH}(x_i))}}) < \LDim(\calH)$ by definition of SOA. Additionally, we have $\LDim({\calH|_{(x_1,\SOA_\calH(x_1)),\dots, (x_k, \SOA_\calH(x_k))}}) < \LDim(\calH)$ by reducibility. Hence, we can take $\calH_i = {\calH|_{(x_i,1 - \SOA_{\calH}(x_i))}}$ for $i\le k$ and $\calH_{k+1} = \calH|_{(x_1,\SOA_\calH(x_1)),\dots, (x_k, \SOA_\calH(x_k))}$. It is easily seen that the union of $\calH_i$'s covers $\calH$ and each $\calH_i$ has a strictly smaller Littlestone dimension.

This interpretation allows us to envision the following: Imagine a scenario where $\calH$ and all its restriction classes are $k$-reducible. Denote $d = \LDim(\calH)$. Then, we can cover $\calH$ by $(k+1)$ hypothesis class of dimension $d-1$. Furthermore, we can cover these sub-classes by even smaller classes of dimension $d-2$. Iterating this, we obtain that we can cover $\calH$ by the union of $(k+1)^d$ classes of Littlestone dimension $0$: namely, $(k+1)^d$ singleton classes. Next, for any learning task of interest, we can just work with these singleton classes and apply standard techniques.

Unfortunately, assuming that $\calH$ and all its restrictions are $k$-reducible is way too strong: the above argument effectively proved that any such class must be finite: in fact, there can be at most $(k+1)^d$ hypotheses in it. Nevertheless, we will explore the preliminary idea in the next subsection, and build up an algorithmic framework that will be used in both the PAC and online learning tasks.

\subsection{Decomposition Dimension}

Trying to formulate the decomposition idea sketched in the previous subsection, we consider the following process to decompose a hypothesis class.

\begin{definition}\label{def: p-decomposition}
Let $p,d\in \mathbb{N}$. Let $\calH$ be a hypothesis class with $\LDim(\calH)\le d$. A $(p,d)$-decomposition of $\calH$ is a (not necessarily complete) binary tree $T$ with the following properties.
\begin{itemize}
    \item Every internal node of $T$ is labeled by a point $x\in \calX$ with two outgoing edges labeled by $(x,0)$ and $(x,1)$.
    \item Every node $v$ of depth $k$ is naturally associated with a sequence of examples: namely $S_v = \{(x_1,y_1),\dots, (x_k, y_k)\}$. We also denote $\calH_v = \calH|_{(x_1,y_1),\dots, (x_k,y_k)}$. 
    \item $T$ is called \emph{valid}, if for every node $v$, we have $\mathrm{depth}(v) \le p\cdot (2^{d-\LDim(\calH_v) + 1} - 1)$, and for every leaf $\ell$, we have that $\mathrm{depth}(\ell) \le p\cdot (2^{d-\LDim(\calH_\ell)} - 1)$ and that $\calH_{\ell}$ is $(p\cdot 2^{d-\LDim(\calH_\ell)})$-irreducible.
\end{itemize}
The degree of $T$ is the largest of $\LDim(\calH_{\ell})$ over all leaves $\ell$. The $(p,d)$-decomposition dimension of $\calH$ is then defined as
\begin{align*}
    \DDim_{p,d}(\calH) := \min_{T: \text{$T$ is valid}} \{ \mathrm{Degree}(T) \} .
\end{align*}
We call any $T$ attaining the minimum degree an \emph{optimal} $(p,d)$-decomposition tree for $\calH$, and call the induced decomposition $\{ \calH_{\ell} \}_{\ell: \text{leaves of $T$}}$ an optimal decomposition. Note that optimal decompositions are not necessarily unique.
\end{definition}

Intuitively, we decompose $\calH$ by selectively choosing points $x$ and split $\calH$ according to the value of $h(x)$ for $h\in \calH$. We want the decomposition to make some ``progress'' in simplifying the problem, in the sense that, for roughly every $p\cdot (2^t-1)$ pairs of restriction, the restricted hypothesis class has its Littlestone dimension reduced by $t$. We also want the tree to be locally maximal in that every leaf is irreducible and the decomposition cannot continue further.

\paragraph*{Basic observations.} To develop some intuition about the definition, we make a couple of simple but important observations here. First, we assert the existence of a valid $p$-decomposition.

\begin{claim}\label{claim: exist decomposition}
    For $p,d,\mathcal{H}$ as in \Cref{def: p-decomposition}, a valid $(p,d)$-decomposition of $\mathcal{H}$ always exists.
\end{claim}

\begin{proof}
    We construct a valid decomposition by using a greedy approach. Let $T$ be the candidate decomposition tree. Initialize $T$ with only a root node (which is also understood as a leaf). Then, whenever there is a leaf $\ell$ whose associated class $\mathcal{H}_\ell$ is $k\le p\cdot 2^{d-\LDim(\mathcal{H}_v)}$-reducible for some $k$, we take $(x_1,\dots, x_{k})$ to be a sequence of inputs that witness the reducibility, and we make a \emph{restriction path} given by $(x_1,\dots, x_k)$: we first make $\ell$ an internal node with two children $\mathcal{H}_\ell|_{(x_1,0)}$ and $\mathcal{H}_{\ell}|_{(x_1,1)}$, and proceed to the branch of $(x_1, \mathrm{SOA}_{\mathcal{H}|_{\ell}}(x_1))$ and recursively restrict $x_2, x_3$ and so on.

    This greedy procedure obviously halts in finite iterations. We now argue the depth requirement is respected. Consider any root-to-leaf path along the resulting tree. We see that it takes at most $p$ restrictions to reduce the Littlestone dimension from $d$ to $d-1$, and $p\cdot 2$ restrictions to reduce dimension from $d-1$ to $d-2$, so on and so forth. We may then conclude that for any internal node $v$ with dimension $d'$, the depth of $v$ is bounded by
    \begin{align*}
        \left( \sum_{i=1}^{d-d'} p\cdot 2^{i-1} \right) + p\cdot 2^{d-d'} - 1 \le p\cdot (2^{d-d' + 1} - 1),
    \end{align*}
    Here, the summation term is the number of steps it takes to reduce the dimension to $d'$. We have the additional term because $v$ may be in a restriction path to further reduce the dimension. Likewise, for a leaf $\ell$ we know that it cannot be in the process of a ``restriction path''. Therefore we only need to count the summation term and get the desired depth upper bound.
\end{proof}

\paragraph*{Bounding the number of leaves.} Next, we will upper-bound the number of leaves produced by a decomposition tree.

\begin{lemma}\label{lemma: number of leaves}
    Let $\calH$ be a hypothesis class of Littlestone dimension at most $d$. Let $T$ be a valid $(p,d)$-decomposition tree of $\calH$. Then, the number of leaves of $T$ is at most $p^d\cdot 2^{d^2}$.
\end{lemma}

\begin{proof}
    We use a potential argument. For every node $u$ of the tree, define the potential of $u$ as $\Phi(u) = (p\cdot 2^{d} - \mathrm{depth}(u))^{\LDim(\calH_u)}$.

    Let $v_1,v_2$ be two children of $u$. Write $t = \mathrm{depth}(u)$ and $\ell = \LDim(\calH_u)$. We observe that
    \begin{align*}
        \Phi(v_1)+\Phi(v_2) \le (p2^{d} - t - 1)^{\ell} + (p2^{d}- t - 1)^{\ell-1} \le (p2^{d} - t)^\ell = \Phi(v).
    \end{align*}
    Thus the sum of potential from two children is no larger than the potential of $u$ itself. By induction, the sum of potential from all leaves is bounded by $\Phi(root) \le p^d 2^{d^2}$. On the other hand, every leaf has potential at least $1$ (because the tree has depth bounded by $p2^{d}-1$). We conclude that the number of leaves is at most $p^d\cdot 2^{d^2}$.
\end{proof}

\paragraph*{Expressiveness of SOA Concepts.} We also need the following lemma from \cite{ghazi2021private-learning}. We refer the readers to \cite{ghazi2021private-learning} for its proof.

\begin{lemma}[Lemma 4.4 in \cite{ghazi2021private-learning}]\label{lemma: Ldim-of-SOA}
For a class $\calH$ with $\LDim(\calH) \le d$, define
\begin{align*}
\hat{\calH}_{d+1} = \{ \SOA_{\calG}: \calG\subseteq \calH \text{ is $(d+1)$-irreducible. } \}.
\end{align*}
Then, it holds that $\LDim(\hat{\calH}_{d+1}) \le d$ as well.
\end{lemma}

\paragraph*{The key lemma.} In Definition~\ref{def: p-decomposition}, the choice of $p\cdot(2^t-1)$ may seem a bit arbitrary. The next lemma, which is the key to the algorithm, will clarify the design.

\begin{lemma}\label{lemma: p-decomposition}
Let $\calG\subseteq \calH$ be two hypothesis classes of Littlestone dimension at most $d$. Let $\{\calG_v\}$ and $\{\calH_{u}\}$ be their optimal $(2p,d)$ and $(p,d)$-decompositions (arbitrarily chosen), respectively. Then, the following statements are true.
\begin{itemize}
    \item $\DDim_{2p,d}(\calG) \le \DDim_{p,d}(\calH)$.
    \item Suppose $\DDim_{2p,d}(\calG) = \DDim_{p,d}(\calH) = t$. Then for every $\calG_v$ with $\LDim(\calG_v) = t$, there exists $\calH_u$ such that $\LDim(\calH_u) = t$ and $\SOA_{\calH_u} = \SOA_{\calG_v}$.
\end{itemize}
\end{lemma}

\begin{proof}
    For the first claim, simply note that the optimal decomposition tree $T_{\calH}$ for $\calH$ gives a candidate decomposition tree for $\calG$. The validity of $T_{\calH}$ w.r.t.~$\calG$ may fail due to possible reducible leaves. Nonetheless, one can always apply a restriction path to reducible leaves, to further reduce the Littlestone dimension of the decomposed sub-classes. Because we had set weaker depth requirement for internal nodes, the depth requirement is obeyed in the process (see also the proof of \Cref{claim: exist decomposition}).  Overall, one can modify $T_{\calH}$ to obtain a valid decomposition tree for $\calG$, which would then witness the upper bound on $\DDim_d(\calG)$.
    
    We establish the second claim here. Suppose for contradiction that the statement fails at some $\calG_v$. We show that $\calG_v$ is $(2p\cdot 2^{d-t})$-reducible, which would invalidate the decomposition $\{\calG_v\}$ according to Item 3 of Definition~\ref{def: p-decomposition}. Let $T_\calH$ be a decomposition tree of $\calH$. Consider running $\SOA_{\calG_v}$ on $T_\calH$. That is, we start at the root of $T_{\calH}$, at every node $v$ with label $x_v$, we proceed to the child with edge $(x_v, \SOA_{\calG_v}(x_v))$. We keep walking on the tree until we have made $2p\cdot 2^{d-t}$ steps, or until we reach a leaf, whichever happens sooner. Let $u$ be the node we end up being at, and let $S_u$ be the sequence of edges traversed. Depending on whether $u$ is an internal node or a leaf, we argue:
    \begin{itemize}
        \item \textbf{Case 1.} If $u$ is an internal node, we know the depth of $u$ is $k' = 2p2^{d-t} = p\cdot 2^{d-t+1} > p (2^{d-t+1}-1)$. Since $T_{\calH}$ is a valid tree for $\calH$, it follows that $\LDim(\calH_u) \le t-1$ and consequently $\LDim(\calG_v|_{S_{u}}) \le t-1$. This shows that $\calG_v$ is $2p2^{d-t}$-reducible, as claimed.
        \item \textbf{Case 2.} Now consider the case that $u$ is a leaf. Because of the bound $\DDim_{p,d}(\mathcal{H})=t$, we have $\LDim(\mathcal{H}_u) \le t$.
        If $\LDim(\calG_v|_{S_u}) \le t-1$, this still means that $\calG_v$ is $k'$-reducible, which leads to the contradiction. Therefore,
        \begin{align*}
            t \ge \LDim(\calH_u) \ge \LDim(\calG_v|_{S_u}) \ge t,
        \end{align*}
        and so we have $\LDim(\calH_u) = t$. Now we claim $\SOA_{\calH_u} = \SOA_{\calG_v}$. Suppose otherwise: namely we have $\SOA_{\calH_u}(x^*) \ne \SOA_{\calG_v}(x^*)$ at some $x^*$. It would follow that
        \begin{align*}
            \LDim(\calG_v|_{S_u\cup \{(x^*, \SOA_{\calG_v}(x^*))\}}) \le \LDim(\calH_u) - 1 \le t-1,
        \end{align*}
        which again implies that $\calG_{v}$ is $2p2^{d-t}$-reducible (recall that the depth of $\mathcal{H}_u$, namely $|S_u|$, is much less than $2p2^{d-t}$, meaning that we can afford one additional restriction here). Hence, we may conclude that $\SOA_{\calH_u} = \SOA_{\calG_v}$, as desired.
    \end{itemize}
    To wrap up, we have proved Item 2 of Lemma~\ref{lemma: p-decomposition}. Our proof even provides an algorithm to find the target $\calH_u$ given $\calG_v$: simply use $\SOA_{\calG_v}$ to traverse the decomposition tree of $\calH$ and stop at the leaf found.
\end{proof}

\paragraph*{Digest.} Lemma~\ref{lemma: p-decomposition} is useful in the following situation: suppose $\mathcal{H}_1,\dots, \mathcal{H}_k$ are $k$ hypothesis classes with the same $(p,d)$-decomposition dimension $t$. Consider their decomposition trees and the SOA hypotheses associated with those dimension-$t$ leaves. If there is a hypothesis $\hat{h}$ that arises as a common SOA in all $k$ trees, one can identify it with a private selection algorithm (see Algorithm~\ref{algo:private sample}), provided that $k$ is moderately large.

What if such a $\hat{h}$ does not exist? We take $\mathcal{G}$ to be the intersection of $\mathcal{H}_1,\dots, \mathcal{H}_k$. Remarkably, Lemma~\ref{lemma: p-decomposition} then implies that $\DDim_{2p,d}(\mathcal{G}) \le t-1$! Indeed, if the $(2p,d)$-decomposition dimension of $\mathcal{G}$ was also $t$, there would be a dimension-$t$ leaf in the decomposition tree of $\mathcal{G}$, whose associated SOA hypothesis should have also appeared in all the $k$ decomposition trees for $\mathcal{H}_1,\dots, \mathcal{H}_k$ (by Item 2 of Lemma~\ref{lemma: p-decomposition}). Hence, we find ourselves in a win-win situation: either $\mathcal{H}_1,\dots, \mathcal{H}_k$ agree on some common hypothesis, or else we make progress by restricting to their intersection.

\paragraph*{Improving the PAC learner.} We now briefly comment on why we are able to shave off a factor of $d$ for the private PAC learner compared with \cite{ghazi2021private-learning}. Overall, our construction is similar to \cite{ghazi2021private-learning}: we split the data set into some $k$ groups and construct interleaving hypothesis classes $\{H_{i}^j\}_{i\in [k],j\in [d]}$. Then, we use the Sparse Vector Technique (SVT) to privately identify the first $j'\in [d]$ such that the algorithm is able to aggregate a ``consensus'' among $\{H_i^{j'}\}_{i\in [k]}$. In contrast, the previous algorithm of \cite{ghazi2021private-learning} (implicitly) used naive composition over $d$ different trials of $j'\in [d]$, incurring a polynomial blow-up in terms of $d$. Although it is possible that one can work harder on the original formulation of \cite{ghazi2021private-learning} and make similar improvements to their algorithm directly, we believe that our new perspective makes it significantly easier to spot this room of improvement.

The discussion and Lemma~\ref{lemma: p-decomposition} motivate the following definition.

\begin{definition}\label{def: essential hypothesis}
    Let $\calH$ be the class and $t = \DDim_{p,d}(\calH)$. A hypothesis $f:\calX\to \{0,1\}$ is called $(p,d)$-essential to $\calH$, if it appears in \emph{every} optimal $(p,d)$-decomposition of $\calH$. Formally, for every optimal $(p,d)$-decomposition $\{\calH_{\ell}\}$ of $\calH$, there exists $\ell$ such that $\LDim(\calH_\ell) = t$ and $\SOA_{\calH_\ell} \equiv f$.
\end{definition}

The following is a direct corollary of Lemma~\ref{lemma: number of leaves} and Lemma~\ref{lemma: p-decomposition}.

\begin{corollary}\label{coro: essential hypotheses}
    Let $\calH$ be the class and $t = \DDim_{p,d}(\calH)$. The following are true:
    \begin{itemize}
        \item There are at most $p^d2^{d^2}$ $(p,d)$-essential hypotheses to $\calH$.
        \item If $\mathcal{G}\subseteq \mathcal{H}$ are two classes with the same $(p,d)$-decomposition dimension, then all $(p,d)$-essential hypotheses of $\mathcal{G}$ are also $(p,d)$-essential of $\mathcal{H}$.
        \item If $\DDim_{2p,d}(\calH) = \DDim_{p,d}(\calH) = t$, there is at least one $(p,d)$-essential hypothesis to $\calH$.
        \item If $t = 0$, then $\mathcal{H}$ is finite, has $|\mathcal{H}|$ essential hypotheses, which are exactly all hypotheses in $\mathcal{H}$.
    \end{itemize}
\end{corollary}

\begin{proof}
    Item 1 follows from Lemma~\ref{lemma: number of leaves} directly.
    
    For Item 2, suppose a hypothesis $f$ is not essential to $\mathcal{H}$. We may take $T$ to be a decomposition tree of $\mathcal{H}$ that ``avoids'' $f$. We then understand $T$ as a candidate decomposition tree of $\mathcal{G}$, extend it as appropriate, to obtain an optimal decomposition tree which avoids $f$ as well.
    
    To see Item 3, let $\calH_v$ be arbitrarily chosen from a $(2p,d)$-optimal decomposition of $\calH$ such that $\LDim(\calH_v) = t$. By Item 2 of Lemma~\ref{lemma: p-decomposition}, $\SOA_{\calH_v}$ must appear in \emph{every} optimal $(p,d)$-decomposition of $\calH$, as desired.

    Finally, the last item holds by definition together with the observation that a class $\mathcal{G}$ with $\LDim(\mathcal{G}) = 0$ must be a singleton class.
\end{proof}

\section{DP-ERM for Littlestone Classes}\label{sec: pac learning}

In this section, we present the improved private PAC learning algorithm. We do so by designing a private empirical risk minimization (ERM) procedure for Littlestone classes. Given the machinery developed in Section~\ref{sec: decomposition}, the algorithm and its analysis fit nicely into one page.

\begin{theorem}\label{thm: DP-ERM-littlestone}
    Let $\calH$ be a class of Littlestone dimension at most $d$. For any $\eps,\delta,\alpha > 0$, there is a bound $n = \widetilde{O}\left( \frac{d^5\cdot \log(1/\delta)}{\alpha \eps} \right)$ and an $(\eps,\delta)$-DP algorithm $\calA$ such that the following is true: given a \emph{realizable} data set of $n$ examples $S=\{(x_1,y_1),\dots, (x_n, y_n)\}$, with probability one $\calA(S)$ outputs a hypothesis $h$ such that $\frac{1}{n}\sum_{i=1}^{n} \mathbf{1}\{ h(x_i)\ne y_i \} \le \alpha$. Furthermore, $h$ is of the form $\SOA_{\calG}(h)$ for some $\calG\subseteq \calH$ that is $(d+1)$-irreducible.
\end{theorem}

Combining Theorem~\ref{thm: DP-ERM-littlestone} with Lemma~\ref{lemma: Ldim-of-SOA} and the standard VC generalization argument, we arrive at the following corollary.

\begin{corollary}\label{coro: PAC learning}
Let $\calH$ be a class of Littlestone dimension $d$. For any $\eps,\delta, \alpha, \beta$, there is an $(\eps,\delta)$-DP algorithm that PAC-learns $\calH$ up to error $\alpha$ with confidence $1-\beta$, using $\widetilde{O}\left( \frac{\log(1/\delta) d^5}{\eps \alpha} \right)$ examples.
\end{corollary}

We now prove Theorem~\ref{thm: DP-ERM-littlestone}.

\begin{proof}[Proof of Theorem~\ref{thm: DP-ERM-littlestone}]
Let $S = \{(x_1,y_1),\dots, (x_n,y_n)\}$ be the data set. Choose $k = \frac{d^2\log(1/\delta)}{\eps}$. We randomly partition the input into $k$ chunks, each of size $\frac{n}{k} \approx \frac{d^3}{\alpha}$. Denote these data sets by $S_1,\dots, S_k$. Let $\alpha = \frac{d^5 \log(1/\delta)}{\eps N}$. 

Consider the following event $E_{good}$: 
\begin{align*}
    \forall h\in \calH, \forall i\in [k], ~~~\err_{S_i}(h) ~
    \begin{cases}
    \in (1\pm \frac{1}{5d})\cdot  \err_S(h) & \text{if $\err_S(h) > \frac{\alpha}{3}$} \\
    \in [0, \alpha/2] & \text{if $\err_S(h) \le \frac{\alpha}{3}$}
    \end{cases}.
\end{align*}
By a standard uniform convergence argument, $E_{good}$ holds with probability at least $0.99$. Alternatively, by Sauer's lemma, we only need to union bound over $|S|^{d}$ different $h$'s. For every fixed $h$, the statement holds with probability $1-2^{-d\log(n)}$ by Proposition~\ref{prop:chernoff-sample-without-replacement}. 

For every $i\in [k]$ and $j\in [d]$, we may define
\begin{align*}
    H_{i}^j = \{h\in \calH : \err_{S_i}(h) \le (1-\frac{1}{2d})^{j} \alpha \}.
\end{align*}
Assuming $E_{good}$, we have for every $i,j$ that
\begin{align*}
    H_{i}^{j+1}\subseteq \bigcap_{i'\in [k]} H_{i'}^j.
\end{align*}
This means we can apply Lemma~\ref{lemma: p-decomposition} between $H_{i}^{j+1}$ and $H_{i'}^j$ for every $(i,i',j)$.

\paragraph*{The algorithm.} We design our algorithm below. The algorithm operates in $d+1$ stages. For the $j$-th stage, $1\le j\le d+1$:
\begin{itemize}
    \item Let $p_j = 2^jnd$. For every $i\in [k]$, let $\{f^{i,j}_v\}_v$ be the $(p_j,d)$-essential hypotheses to $H_i^{j}$.
    \item Use AboveThreshold with privacy parameter $\eps$ to test if there is a common hypothesis $\hat{h}$ that appears at least $\frac{k}{2}$ times among the list $\{f^{i,j}_v\}_{i,j,v}$. Note that this is asking for the frequency of ``the most frequent item'' in a list, which is a sensitivity-$1$ query and fits in the template of AboveThreshold.
    \begin{itemize}
        \item If the answer is yes, use Algorithm~\ref{algo:private sample} (Sparse Sample) to output one such $\hat{h}$ and halt.
        \item Otherwise, continue to the $(j+1)$-th stage of the algorithm.
    \end{itemize}
\end{itemize}
This completes the design of the algorithm. By the privacy property of AboveThreshold and Sparse Selection, the algorithm is easily seen to be $(\eps,\delta)$-DP.

\paragraph*{Utility analysis.} We first show that, with high probability, the algorithm outputs a hypothesis before the $(d+1)$-th stage concludes. To see this, let us track how $\max_i\{\DDim_{p_j,d}(H_i^j)\}$ evolves as $j$ increases. if the algorithm fails to find a common hypothesis $\hat{h}$ in the $j$-th round, it must be the case that $\max_i\{\DDim_{p_{j+1},d}(H_i^{j+1})\}\le \max_i\{\DDim_{p_j,d}(H_i^j)\} - 1$ by Lemma~\ref{lemma: p-decomposition}. Since we start with $\DDim_{p_1,d}(H_i^1) \le d$, the decrease can only happen at most $d$ times. So the algorithm must be able to find a common hypothesis during one stage (this analysis assumes $E_{good}$ and the event that the AboveThreshold and SparseSelection algorithms both behave reasonably, which happens with high probability by standard arguments).

We next prove that the hypothesis $\hat{h}$ produced by the algorithm has low empirical error. Indeed, by design of the algorithm we know that $\hat{h} = \SOA_{\calG}$ for some $\calG\subseteq H_i^j$ that is $nd$-irreducible. Suppose $\err_{S}(\hat{h}) > 2\alpha$. Recall the definition of $H_i^j$: it only consists of hypotheses $h$ such that $\err_S(h) \le 2\alpha$. This consequently implies that $\calG|_{(x_1,\hat{h}(x_1)),\dots, (x_n,\hat{h}(x_n))} = \emptyset$: simply because no hypothesis $h$ in $\calG$ can behave like $\hat{h}$ and make so many mistakes on $x_1,\dots, x_n$! As this is a contradiction to the irreducibility of $\calG$, our assumption must be false and it is the case that $\err_{S}(\hat{h}) \le 2\alpha$, completing the analysis.

Overall, we have shown an algorithm that works with high constant probability. By known tricks, this can be turned into an algorithm with success probability one, without affecting the asymptotic bounds.
\end{proof}

\begin{remark}
    As a final remark to our algorithm, we note that one can use the improper-to-proper transformation of \cite{ghazi2021private-learning} to make our algorithm proper (i.e., design an algorithm that always outputs some $h\in \mathcal{H}$ with low empirical error). The only property we need for this transformation is that our algorithm as described above always outputs a hypothesis $h = \mathrm{SOA}_{\mathcal{G}}$ where $\mathcal{G}\subseteq \mathcal{H}$ is $d$-irreducible. We refer the readers to \cite{ghazi2021private-learning} for more details.
\end{remark}

\section{Private Online Learning in the Realizable Case}\label{sec: online realizable}

In this section, we design our algorithm for private online learning of Littlestone classes with mistake bounded by $\poly(d, \log(T))$. Given the known lower bound of $\Omega(d + \log(T))$ for this task, our algorithm characterizes the mistake bound of private online learning up to polynomial factors.



\paragraph*{Algorithm overview.} We (very briefly) sketch the algorithm idea. We will follow the subsample-and-aggregate framework as in Section~\ref{sec: pac learning}. For some $k$ to be chosen, we wish to train $k$ teachers and let them agree on some common hypothesis $\hat{h}$, allowing for the publication of $\hat{h}$ privately. We then use $\hat{h}$ to answer the online queries to come. Once we have gathered enough (of order $\poly(d)$) ``counter examples'' to $\hat{h}$, we will randomly distribute the examples to $k$ teachers and update them. By carefully designing the update strategy, we can bound the number of update rounds by $\poly(d)$. This and other reasons dictate the choice of $k$ to be $\frac{1}{\eps}\poly(d,\log(T))$. Overall, we can ensure a mistake bound of $\poly(d, \log(T))$ with high probability.

We present the algorithm in its pseudo-code, shown in Algorithm~\ref{algo:private online learning}.

\begin{algorithm}[h]
\LinesNumbered
    \caption{Private Online Learning (DP-OL)}
    \label{algo:private online learning}
    
    \SetKwProg{Fn}{Function}{:}{}
    
    \DontPrintSemicolon
    
    \SetKwFunction{OnlineLearn}{PrivateOnlineLearn}
    \SetKwFunction{JointTrain}{JointTrain}
    \SetKwFunction{DefineClass}{DefineClass}
    \SetKwFunction{GrowTree}{GrowTree}
    \SetKwComment{Comment}{/* }{ */}
    
    \KwIn{
         Hypothesis class $\calH$ with $d = \LDim(\calH)$; Privacy parameters $\eps,\delta > 0$; Total time steps $T\ge 1$.
    }
    
    \Fn{\OnlineLearn{}}{
         $K_{\mathrm{budget}}\gets C \cdot d^{3}$ \Comment*[r]{Privacy budget control counter}
         $k\gets \frac{d^{3.5} \log(T) \log(1/\delta) }{\varepsilon}$ \Comment*[r]{The number of `teachers'.}
         $U\gets C\cdot d^3 \cdot (\log(d\log(T)))\cdot k$ \Comment*[r]{Re-train after every $\approx U$ mistakes.}
         $S^{(1)},\dots, S^{(k)}\gets \emptyset$ \Comment*[r]{Each $S^{(i)}$ maintains a \emph{collection} of data sets.}
         $\hat{h}\gets \JointTrain(k, S^{(1)},\dots, S^{(k)})$ \;
             $E\gets \emptyset$ \Comment*[r]{a buffer set to store mistake examples.}
        \For{$t=1,\dots, T$}{
            Receive query $x_t$ and Predict $\hat{h}(x_t)$ by \emph{publishing} $\hat{h}$ \;
            Receive the correct label $y_t$ \;
            \If{$y_t \ne \hat{h}(x_t)$}{
                Algorithm has made one mistake \;
                $E\gets E\cup \{(x_t, y_t) \}$ \;
            }
            \If(\tcp*[f]{AboveThreshold Test}){$|E| + \mathrm{Lap}(\frac{\log(T/\delta) d^3}{\eps}) > U$}{
                $K_{\mathrm{budget}} \gets K_{\mathrm{budget}} - 1$ \;
                \If{$K_{\mathrm{budget}} < 0$}{
                    HALT the algorithm \;
                }
                Randomly split $E$ into $A_1,\dots, A_k$ of equal size \;
                $S^{(i)} \gets S^{(i)} \cup \{ A_i \}$ for every $i$ \;
                $\hat{h} \gets \JointTrain(k, S^{(1)},\dots, S^{(k)})$ \Comment*[r]{Refer to Algorithm~\ref{algo: joint train}}
                $E\gets \emptyset$ \;
            }
        }
    }
\end{algorithm}

\begin{algorithm}[h]
\LinesNumbered
\DontPrintSemicolon
    \caption{Jointly train $k$ teachers}
    \label{algo: joint train}
    
    \SetKwProg{Fn}{Function}{:}{}
    \SetKwFunction{SparseSample}{SparseSample}
    \SetKwComment{Comment}{/* }{ */}

    \KwIn{
         $k$ collections of data sets $S^{(1)},\dots, s^{(k)}$ \;
    }
    \SetKwInput{GValue}{Global Variable}
    
    \GValue{A stage indicator $j^*\ge 1$, initialized to $1$\;}
    \KwOut{
        A hypothesis $\hat{h}:\calX\to \{0, 1\}$ \;
    }
    \Fn{\JointTrain{$k, S^{(1)},\dots, S^{(k)}$}}{
        \While{$j^* \le d+1$}{
            \For{$i=1,\dots, k$}{
                $H_i^{{j^*}} \gets \DefineClass(\calH, S^{(i)}, j^*)$ \;
                $\calL_i^{j^*} \gets $ the list of $(2^{j^*} d^3, d)$-essential hypotheses to $H_i^{j^*}$ \;
            }
            $m^{j^*}\gets $ the maximum frequency of a hypothesis in $\calL_1^{j^*}\cup \dots \cup \calL_k^{j^*}$ \;
            \eIf{$m^{j^*} + \mathrm{Lap}(O(\log(T/\delta) \cdot d/\varepsilon)) > \frac{4k}{5}$}{
                \For{$r=1,\dots, O(\log(T))$}{
                    $h^{(r)} \gets \SparseSample(\calL_1^{j^*},\dots,\calL_k^{j^*}; \eps_{sparse} = \frac{\eps}{\sqrt{d^3}\log(T)\log(1/\delta)}, B = \frac{k}{10})$ \Comment*[r]{See~Algorithm~\ref{algo:private sample}}
                }
                \Return{$\mathrm{Majority}(h^{(1)},\dots,h^{O(\log(T))})$} \;
            }{
                $j^*\gets j^* + 1$ \;
            }
        }
    }
    \Fn{\DefineClass{$\calH, S, j$}}{
        Parse $S = \{T_1, T_2,\dots, T_\ell\}$ where each $T_i$ is a sub-dataset. \;
        \Return{$H := \{ h\in \calH: \forall i\in [\ell], \err_{T_i}(h) \le \frac{1}{10}\cdot  (1-\frac{1}{d})^j\}$}
    }   
\end{algorithm}

\subsection{Privacy Analysis}

The privacy analysis of Algorithm~\ref{algo:private online learning} is rather modular. Let $K$ be the number of calls to Algorithm~\ref{algo: joint train} from Algorithm~\ref{algo:private online learning}. We think of $K$ as a piece of public information and halt the algorithm whenever $K \ge K_{\mathrm{budget}} = Cd^3$ for a large constant $C$. We will understand too many calls to Algorithm~\ref{algo: joint train} as a utility failure. With this in mind, the following calculation is based on the assumption that $K < O(d^3)$. We list all privacy-leaking components of Algorithm~\ref{algo:private online learning} in the following.
\begin{itemize}
    \item On Line 15 of Algorithm~\ref{algo:private online learning} and Line 8 of Algorithm~\ref{algo: joint train}, we used the AboveThreshold Test, conveniently implemented in the Cohen-Lyu style. We use the template of \Cref{algo: abovethreshold} to count every positive outcome of Line~15, with a sensitivity-to-noise ratio as $\eps' \le \frac{\eps}{d^3\log(T/\delta)}$. We also the same template to count every \emph{negative} outcomes of Line~13 (see the remark following \Cref{lemma: privacy of abovethreshold}), where we set a sensitivity-to-noise ratio as $\eps'' = \frac{\eps}{d\log(T/\delta)}$.

    The number of positive outcomes in the former scenario is bounded by $K = O(d^3)$ the number of negative outcomes in the latter is bounded by $(d+1)$. Overall, we may use \Cref{lemma: privacy of abovethreshold} to conclude that the privacy loss from this part is $(O(\eps_{thr}), O(\delta))$-DP where $\eps_{thr} \le \log(1/\delta)\cdot \eps'\cdot d^3 + \log(1/\delta)\cdot \eps'' \cdot d \le O(\eps)$.
    
    \item Whenever \Cref{algo: joint train} is called, on Line~10 of Algorithm~\ref{algo: joint train}, we will repeatedly call $\SparseSample$ for $O(\log(T))$ times, where each call is $(\eps_2,\hat{\delta})$-DP where $\eps_2 = \frac{\eps}{\sqrt{d^{3}}\log(T)\log(1/\delta)}$ and $\hat{\delta} \le \frac{\delta^2}{d^{10}}$. Since $\hat{\delta}$ is so small, we ignore it in the following calculation.
\end{itemize}

Thus, by advanced composition, given an desired final ``delta'' $\delta$, we can work out the asymptotic ``epsilon'' as
\begin{align*}
        \sqrt{K \log(T) \log(1/\delta)} \eps_2 \le O(\eps),
\end{align*}
provided that $K\le O(d^3)$. Hence, we conclude that the whole algorithm is $(O(\eps),\delta)$-DP as long as $K \le O(d^3)$. We can ensure this is the case, by halting the algorithm once the AboveThreshold test on Line~14 of Algorithm~\ref{algo:private online learning} passes more than $K$ times. Once again, the addition of a hard-stop instruction to Algorithm~\ref{algo:private online learning} does not affect the privacy analysis, because the outcomes of the AboveThreshold tests are publicly available once we account for their privacy cost.

\subsection{Utility Analysis}

We now prove the mistake bound of Algorithm~\ref{algo:private online learning}. Since we deal with oblivious adversaries, let us fix the query sequence $(x_1,y_1),\dots, (x_T, y_T)$, independent of the internal randomness of the algorithm. We start with the following simple claims.

\begin{claim}\label{claim: large U}
    Consider Algorithm~\ref{algo:private online learning}. With probability $1-\frac{1}{T^{10}}$ the following is true: every time the test on Line~14 passes, it holds that $|E| \in (\frac{U}{2}, \frac{3U}{2})$.
\end{claim}

\begin{proof}
Union-bound over the $T$ Laplace noises involved.   
\end{proof}

\begin{claim}\label{claim: interleaving data}
    Consider Algorithm~\ref{algo:private online learning}. With probability $1-\frac{1}{T^{10}}$ the following is true: every time Line~15 is executed, the produced data sets $A_1,\dots, A_k$ satisfy that $\err_{A_i}(h) \in (1\pm \frac{1}{5d}) \err_{A_j}(h)$ for every $i,j\in [k]$ and $h\in \calH$.
\end{claim}

\begin{proof}
    We first condition on the event that $|E|\ge \frac{U}{2}$, which happens with high probability by Claim~\ref{claim: large U}. For every fixed $i,j,h$, the statement is true with probability $1-\exp(-\Omega(|E|/(k d^2))) \ge 1 - |U|^{-2d}$ by the multiplicative Chernoff bound. We then union-bound over all $i,j,h$, noting that there are at most $k^2\cdot |U|^d$ such tuples.
\end{proof}

\begin{claim}\label{claim: interleaving class}
Under the event of Claim~\ref{claim: interleaving data}, the following is true: every time Algorithm~\ref{algo: joint train} is called, the classes $H_i^j$ defined on Line 4 satisfy that $H_i^{j+1} \subseteq H_{i'}^j$ for every $i,i'\in [k]$ and $j\in [d+1]$.
\end{claim}

\begin{proof}
    For every $u\ge 1$, consider the $u$-th call to Algorithm~\ref{algo: joint train}. We know that $S^{(i)},S^{(i')}$ are both collections of $u$ sub-datasets. Write $S^{(i)} = \{ A_i^{1},\dots, A_i^u \}$ and $S^{(i')} = \{ A_{i'}^1,\dots, A_{i'}^{u}\}$. Then, by definition, $h\in H_i^{j+1}$ is equivalent to that $\err_{A_i^{q}}(h) \le \frac{1}{10}(1-\frac{1}{5d})^{j+1}$ for every $1\le q\le u$. Under the event of Claim~\ref{claim: interleaving data}, this implies that $\err_{A_{i'}^q}(h) \le \err_{A_i^{q}}(h)\cdot (1+\frac{1}{5d}) \le \frac{1}{10}(1-\frac{1}{5d})^{j}$ for every $1\le q\le u$, putting the hypothesis into the class $H_i^j$. This completes the proof.
\end{proof}

\subsubsection{Proof of the Mistake Bound}


Before we start the proof, we give a high-level description about what \Cref{algo:private online learning} is doing. We think of the online learning process as having $(d+1)$ stages, one for each $j^*$. For every stage $j^*$, we start with $k$ ``teachers'' with hypothesis classes $H_1^{j^*},\dots, H_k^{j^*}$, each producing as many as $2^{d^2}$ candidates (by \Cref{coro: essential hypotheses}, see Line~5 of \Cref{algo: joint train}). We use $\SparseSample$ to aggregate the candidates and produce a model $\hat{h}$ which is used answer queries until it has made $\approx kd^3$ mistakes. We will prove a key claim next (i.e.,~\Cref{claim: analyze p j}), which intuitively says that, by taking the new mistakes into account, the number of valid candidates shrinks by a constant factor! As such, inside each stage one can iterate this process for at most $O(d^2)$ rounds before running out of candidates. Once a stage is finished, we move into the next stage, where each teacher can again produce as many as $2^{O(d^2)}$ hypotheses. But we still gain from this as the decomposition dimension is provably reduced by at least one: because no common essential hypothesis was left in the previous stage, we can use \Cref{lemma: p-decomposition} to say that, after passing to the intersection of teacher classes (\Cref{claim: interleaving class} and Line 19 or \Cref{algo: joint train}), the decomposition dimension is reduced. As such, we can bound the total number of stages by $(d+1)$. Once the decomposition dimension is reduced to zero, the hindsight consistent hypothesis $h^*$ will show up in the pool of candidates. Once it becomes the only candidate, the algorithm makes no more mistakes.

The following lemma is the first step to confirm the picture we asserted above.

\begin{lemma}\label{lemma: bounded update rounds}
    With probability $1-\frac{1}{T^{10}}$, Algorithm~\ref{algo:private online learning} calls Algorithm~\ref{algo: joint train} for at most ${O}(d^3)$ times.
\end{lemma}
 The rest of the subsection is mostly devoted to the proof of \Cref{lemma: bounded update rounds}. Once the lemma is proved, we will explain how it implies the desired mistake bound at the end. 
 
 With these in mind, for every $u\ge 1$, consider the $u$-th call to the procedure $\JointTrain$. We examine the following quantities:
\begin{itemize} 
    \item The variable $j\in [d+1]$ such that $\JointTrain$ returns with $j^* = j$. We write $j(u)$ to highlight that it is a function of $u$.
    \item For every call to $\JointTrain$, we only enter the Line-9 loop of \Cref{algo: joint train} once. When we enter the loop, let $p(u)$ be the probability of getting $\perp$ from $\SparseSample$.
\end{itemize}
The following claim is the key to our analysis.
\begin{claim}\label{claim: analyze p j}
With probability $1-\frac{1}{T^2}$, all of the following hold: for every $u\ge 1$, we have $\frac{1}{100}\ge p(u)\ge 2^{-O(d^2)}$. Moreover, from $u$ to $u+1$, at least one of the following two things happens:
\begin{itemize}
    \item $j(u+1) \ge j(u) + 1$;
    \item $p(u+1) \ge p(u)\cdot (1+ \frac{1}{10})$.
\end{itemize}
\end{claim}

\begin{proof}
Let us first establish the upper and lower bound for $p(u)$. The upper bound is easy because once the AboveThreshold test on Line 8 of \Cref{algo: joint train} passes, there is at least one candidate with score $\frac{2}{3}k$, while the symbol ``$\perp$'' is only assigned a score of $B = \frac{k}{10}$. An easy calculation reveals that the said candidate is much more likely to be sampled than $\perp$. We turn to the lower bound now. Note that the total number of candidates in ($\mathcal{L}_1 \cup \dots \cup \mathcal{L}_k$) is at most $2^{O(d^2)}$ by \Cref{coro: essential hypotheses}. Hence, even assuming all the $2^{O(d^2)}$ candidates have maximum scores (i.e., $k$), we still get
\begin{align*}
    p(u) = \Pr[\perp \text{ sampled}] \ge \frac{2^{B\eps}}{2^{O(d^2)} \cdot 2^{k\eps}} \ge 2^{-O(d^2)}.
\end{align*}
We turn to the second part of the claim. Note that $j(u)$ is non-decreasing in $u$ by design. So it suffices to prove that, if $j(u+1) = j(u)$, then we have that $p(u+1) \ge p(u) \cdot (1 + \frac{1}{10})$.

Let $D$ be the distribution of $\SparseSample(\mathcal{L}_1,\dots, \mathcal{L}_k)$ during the $u$-th call to \Cref{algo: joint train}. As a quick definitional check, we have $p(u) = \Pr[D = \perp]$. Because of the bound $p(u) \le \frac{1}{100}$, when we draw $O(\log(T))$ samples from $D$, at least $\frac{9}{10}$ fraction of those samples are actual hypotheses instead of $\perp$. Taking $\hat{h}$ to be their majority vote, with probability $1 - \frac{1}{T^{20}}$ over $\hat{h}$, we have for all future queries $(x_{t+1},\dots, x_T)$ that:
\begin{align*}
    |\hat{h}(x_i) - \E_{h\sim D}[h(x_i)]| \le \frac{2}{3}.
\end{align*}
We condition on this event. This means whenever $\hat{h}$ makes a mistake on some $x_i$, at least $\frac{1}{3}$-fraction of hypotheses in $D$ make the same mistake on the query $x_i$. 

Now, let $E$ be the collection of all mistakes made by $\hat{h}$ between the $u$-th and $(u+1)$-th call to the algorithm. For every $(x,y)\in E$ and every $h\in \mathrm{supp}(D)$, we say that $(x,y)$ is a \emph{counterexample} to $h$, if $h(x)\ne y$. We mark a hypothesis $h\in \mathrm{supp}(D)$ as \emph{exposed}, if it gets more than $\frac{1}{6}|E|$ counterexamples out of $E$.

Let $\xi\in [0, 1]$ be the fraction of exposed hypothesis under the measure $D$. For a worst-case analysis, we consider the case that every non-exposed hypothesis has made $\frac{1}{6}|E|$ mistakes on $E$, and every exposed hypothesis has made $|E|$ mistakes. Because on average every hypothesis makes at least $\frac{1}{3} |E|$ mistakes, we get
\begin{align*}
    \xi \cdot 1 + (1-\xi) \cdot \frac{1}{6} \ge \frac{1}{3},
\end{align*}
which translate to $\xi \ge \frac{1}{5}$ (we note that this argument is colloquially known as the ``reverse Markov inequality'').

\paragraph*{Ruling out hypotheses efficiently.} Before continuing further, let us state and prove the following simple but important fact.

\begin{fact}\label{fact: rule out high error}
    Assume $p,d,n$ are such that $p\ge n\ge d$. Suppose $\mathcal{H}$ is a hypothesis class and $f$ is $(p,d)$-essential to $\mathcal{H}$. Let $A\in(\mathcal{X}\times \{0,1\})^n$ be a data set such that $\err_A(f) \ge \frac{1}{8}$. 

    Define $\mathcal{G} = \{h\in \mathcal{H} : \err_A(h) \le \frac{1}{10} \}$. Then, $f$ is not $(p,d)$-essential to $\mathcal{G}$.
\end{fact}

\begin{proof}
    Assume for contradiction that $f$ is essential. Moreover suppose $f = \SOA_{\mathcal{G}_\ell}$ where $\mathcal{G}_{\ell}$ is a leaf in an optimal decomposition tree. Consider the set of inputs $A_x = \{x:(x,y)\in A\}$. Starting with $\mathcal{G}_\ell$ and restricting to $(x, f(x))$ for every $x\in A_x$, we end up with an \emph{empty} hypothesis class by definition of $f$ and $\mathcal{G}$. As we have assumed that $f = \SOA_{\mathcal{G}_\ell}$, this argument implies that $\mathcal{G}_{\ell}$ is not $n$-irreducible, a contradiction. Therefore, such $\mathcal{G}_v$ cannot exist and $f$ is indeed not $(p,d)$-essential to $\mathcal{G}$.
\end{proof}

Bcak to the proof of \Cref{claim: analyze p j}, let $D'$ be the induced distribution of $\SparseSample(\mathcal{L}_1,\dots, \mathcal{L}_k)$ during the $(u+1)$-th call. Since we have assumed that $j(u+1) = j(u)$, both of the following are true: (1) due to Item 2 of \Cref{coro: essential hypotheses}, no new hypothesis is introduced in $D'$ compared with $D$; and (2) applying \Cref{fact: rule out high error} together with \Cref{claim: interleaving data} on each $H_i^{j^*}$ and $A_i$ (c.f.~Line~20 of \Cref{algo:private online learning}), it follows that all the exposed hypotheses are removed from $D'$. As a consequence, we obtain that
\begin{align*}
    p(u+1) = \Pr[D = \perp] \ge (1 + \frac{1}{10}) \Pr[D' = \perp] = (1 + \frac{1}{10}) p(u),
\end{align*}
as claimed.
\end{proof}

\paragraph*{Remark.} We pause here and make a side note. We have tried to implement the proof plan with different private selection protocols (most notably Report-Noisy-Max), but we failed to give a rigorous argument to bound the number of iterations in each stage. We then moved on to the Sparse Exponential Mechanism of \cite{GhaziKM20}, first trying to prove the conclusion by tracking the number of ``candidates'' directly. But then a new complication arises: in particular, each candidate can be supported by a \emph{subset} of teachers, and the subsets are dynamically changing. Finally, it turned out that tracking the probability of sampling the failure symbol ``$\perp$'' gives so far the cleanest and simplest argument. It may be an interesting question to gain a better understanding about the dynamics of the learning process, which can also help design algorithms against an adaptive adversary.

\paragraph*{Completing the proof.} Having Claim~\ref{claim: analyze p j}, the proof of \Cref{lemma: bounded update rounds} is immediate: once in every $O(d^2)$ calls we know $j(u)$ must increase by at least one. Since $j(u)$ is always upper bounded by $d+1$, the lemma is established. We conclude the proof by proving the mistake bound below.

\begin{lemma}\label{lemma: bounded mistakes}
    On a realizable query sequence, with probability $1-\frac{1}{T^{10}}$, Algorithm~\ref{algo:private online learning} makes at most $O(d^3 U) \le \widetilde{O}(\frac{1}{\eps}d^{9.5}\cdot \log(T)\log(1/\delta))$ mistakes.
\end{lemma}

\begin{proof}
    Note that \Cref{claim: large U} implies that the algorithm will call $\JointTrain$ whenever $\frac{3U}{2}$ mistakes are made, while \Cref{lemma: bounded update rounds} says that there are at most $O(d^3)$ calls to $\JointTrain$. Since the sequence is realizable, there is always a hypothesis $h^*$ consistent with all query pairs. As such, the hypotheses class $H_i^{j^*}$ defined in \Cref{algo: joint train} can never be empty and they always have $h^*$ in their intersection. 
    
    Next, by \Cref{claim: interleaving class} and \Cref{lemma: p-decomposition}, whenever $j(u)$ increases by one, the decomposition dimensions of the teacher classes are reduced by at least one. When $j(u)$ increases to the maximum (i.e., $d+1$), the decomposition dimension must have reduced to zero. Item 4 of \Cref{coro: essential hypotheses} then implies that the teacher classes still share $h^*$ as the common candidate, and $\SparseSample$ would not fail.
    
    All in all, with probability at least $1-\frac{1}{T}$, the algorithm does not halt prematurely, and makes at most $\widetilde{O}(\frac{d^{9.5}\cdot \log(T)\log(1/\delta)}{\varepsilon})$ mistakes as claimed.
\end{proof}

\begin{remark}\label{remark:adaptive}
    The most crucial place where we required oblivious adversary is in the proof of Claim~\ref{claim: analyze p j}: in particular, when we claim the uniform approximation of $\hat{h}$ to future queries can be achieved by drawing only $O(\log(T))$ hypotheses from the ``$\mathrm{SparseSample}$'' oracle. If the adversary is adaptive, it can first observe $\hat{h}$ and come up with queries for which $\hat{h}$ is not representative of the underlying distribution. Nevertheless, by the dual VC dimension bound and a uniform convergence argument, one can draw $2^{O(d)}$ hypotheses and guarantee that their majority vote \emph{uniformly} approximates the evaluation on \emph{every} point! Adding this new ingredient on top of the existing framework seems to allow us to design an algorithm against \emph{adaptive} adversary with a mistake bound of $2^{O(d)}\cdot \mathrm{polylog}(T)$.
    
    However, establishing the result for the adaptive setting does require extra formalism and care (for one, we can no longer conveniently fix the query sequence $(x_1,y_1),\dots, (x_T, y_T)$  beforehand), which is beyond the scope of this paper.
\end{remark}

\section*{Acknowledgement}

I am grateful to Jelani Nelson for useful conversations, to Peng Ye for insightful discussions on topics related to private learning and sanitation. I would also like to thank anonymous reviewers, whose comments and suggestions have significantly improved the presentation.

X.L. is supported by a Google PhD fellowship.

\addcontentsline{toc}{section}{References}
\bibliographystyle{alpha}
\bibliography{submission/references}

\appendix

\end{document}